\documentclass[11pt]{article}
\usepackage[final]{acl}
\usepackage{times}
\usepackage{latexsym}
\usepackage[T1]{fontenc}
\usepackage[utf8]{inputenc}
\usepackage{microtype}
\usepackage{graphicx}
\usepackage{booktabs}
\usepackage{amsmath}
\usepackage{tikz}
\usetikzlibrary{positioning,arrows.meta}
\makeatletter
\def\paragraph{\@startsection{paragraph}{4}{\z@}{0.2ex}{-1em}{\normalsize\bfseries}}
\def\section{\@startsection{section}{1}{\z@}{-1.2ex plus -0.2ex minus -.2ex}{0.6ex plus 0.2ex minus .2ex}{\large\bfseries\raggedright}}
\makeatother
 
\newcommand{\gpt}{GPT-5.6-terra}
\newcommand{\opus}{Claude Opus~5.5}
\newcommand{\haiku}{Claude Haiku~4.5}
 
\title{Do Personalized LLMs Need the Whole Conversation?\\Matching the Full History with Only 5\% of Its Text \\via Relevance-Filtered Context }
 
\author{Erika Hayashi \\
  Keio University\\%, Yokohama, Japan \\
  \texttt{erika0424@keio.jp} \\\And
  Andrew Shin \\
  Keio University\\%, Yokohama, Japan \\
  \texttt{shin@inl.ics.keio.ac.jp}}
  
\begin{document}
\maketitle
 
\begin{abstract}
Personalized assistants accumulate long conversation histories, and passing the whole history to an LLM with every request is costly.
We show that it is also unnecessary.
On PersonaMem-v2 (180 questions), with \gpt{} and \opus{} as judges, a \emph{relevance-filtered context}, built by sentence-level retrieval that ranks the history by similarity to the current request and stops at a fixed budget, matches full-history accuracy with only 5\% of the text, cutting the judge's input about 20-fold.
An LLM summary of the same length performs on par, but it needs an LLM to read the entire history, whereas retrieval needs only a small embedding model.
While both sentence-level retrieval and LLM summary perform strongly, retrieval does so while placing the relevant evidence in the context for only 61\% of the questions, and an oracle given only that evidence beats the full history by 15.6 points, making better context selection a promising path to surpassing the full history.
\end{abstract}
 
\section{Introduction}
Personalized assistants that serve a user over many sessions accumulate conversation histories of tens of thousands of tokens.
A request such as ``suggest a morning routine'' may depend on a preference the user mentioned weeks earlier, so the earlier statement must be made available to the model.
However, the obvious solution, appending the whole history to the prompt, has three drawbacks: it is expensive, it hits context limits as histories grow, and it does not guarantee accuracy, because language models use information in long inputs unevenly \citep{liu2024lost} and often fall short of their advertised context length \citep{hsieh2024ruler}.
 
Practitioners therefore shorten the history in three main ways: truncating it, summarizing it \citep{wang2023recursive,zhong2024memorybank,jiang2023llmlingua}, or retrieving from it \citep{lewis2020rag,packer2023memgpt,park2023generative}.
However, determining which of these approaches works best can be misleading for three reasons.
First, methods are often compared with different amounts of text, which confounds the method with the budget.
Second, comparisons rarely include references that separate ``the history is not needed'' cases from ``the context lost the relevant information'' cases. In fact, for questions we generated ourselves from conversations, both judges answered about half correctly with no conversation at all, against a chance level of 25\%. Third, summaries are known to drop details such as numbers, yet in our earlier experiments such effects could not be separated from how answerable the questions were without the history and from differences in the lengths compared (Appendix~\ref{app:earlier}).
 
\begin{figure*}[t]
\centering
\includegraphics[width=0.97\textwidth,trim=0 25 0 22,clip]{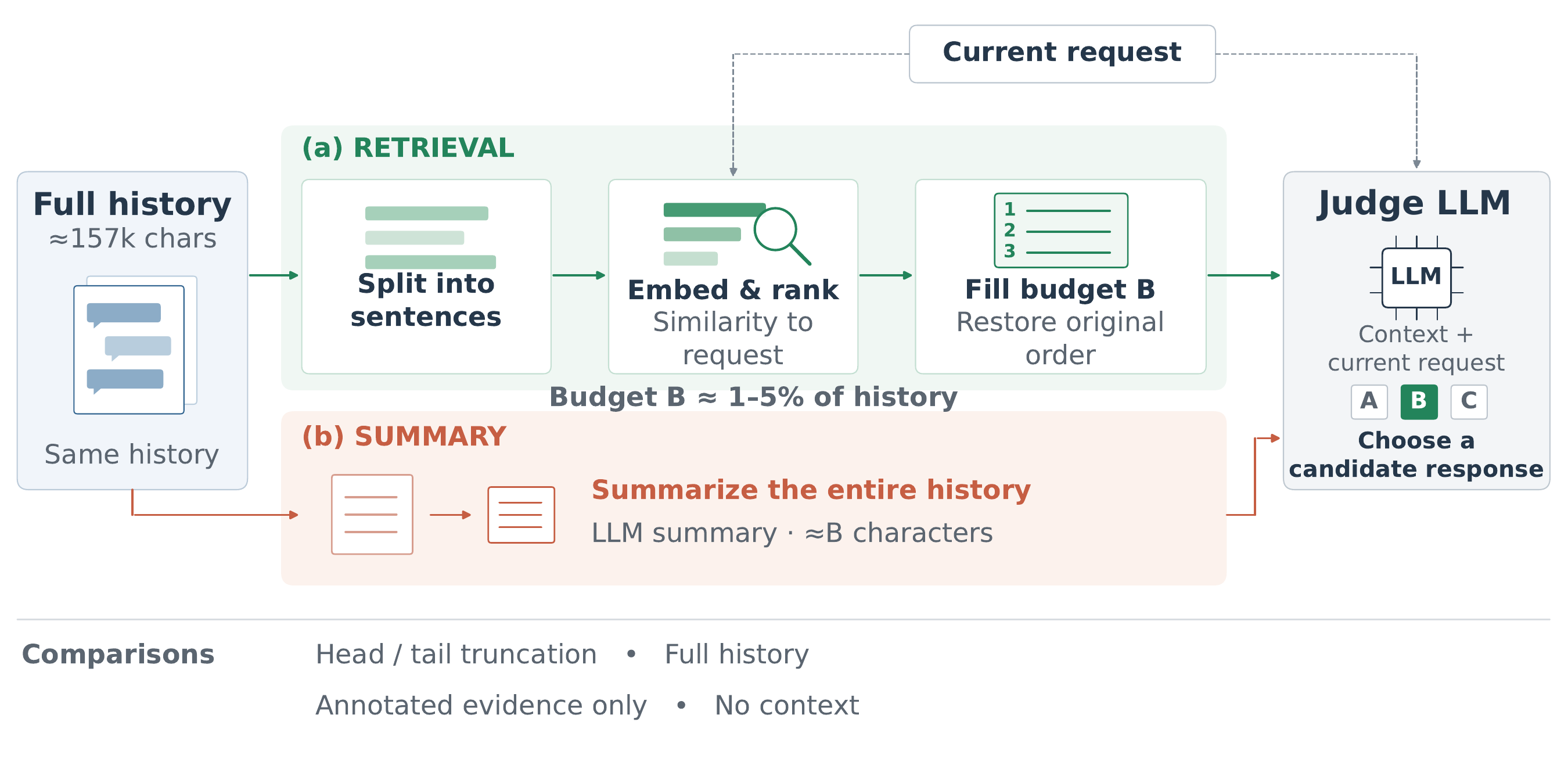}
\caption{Overview of the compared contexts. The history is
turned into a compact context of $B$ characters by
(a) sentence-level retrieval of parts relevant to the
current request (\textit{relevance-filtered context}) or (b) an LLM summary of the whole history.}
\label{fig:pipeline}
\end{figure*}

We therefore examine a \emph{relevance-filtered context} (Figure~\ref{fig:pipeline}): split the history into sentences, rank them by similarity to the user's current request, fill a small character budget with the top-ranked sentences, and restore their original order. Because the relevant part of the history is unknown until a request arrives, the practical baseline is the full history, and we ask how small a relevance-filtered context can be while still matching it. We compare it with LLM summaries of the same size, truncation, the full history, and two references (no context, and an oracle that sees the \textit{evidence only}), on the PersonaMem-v2 text benchmark \citep{jiang2025personamemv2} with two judges, \gpt{} and \opus{}.
Our main findings are: (i) a relevance-filtered context of 5\% of the history is on par with the full history for both judges, so the context can stop there, whereas truncated histories, which lack relevance ordering, keep improving up to their full length;
(ii) an LLM summary of the same size performs on par, but retrieval needs no LLM call and updates incrementally as the history grows;
(iii) the \textit{evidence only} oracle beats the full history by 15.6 points, while retrieval still misses the evidence for 39\% of the questions, which makes context selection the lever for surpassing the full history.
 
\section{Related Work}
\paragraph{Using long inputs.}
Long-context models still use information unevenly: accuracy depends on where relevant text appears \citep{liu2024lost}, and many models do not sustain their claimed context length on synthetic tasks \citep{hsieh2024ruler}.
These studies motivate shortening the input, but they do not compare ways of doing so on personalization queries at equal budget.
 
\paragraph{Summarization and compression.}
Summaries and compressed prompts reduce the input length: recursive summarization of dialogues \citep{wang2023recursive}, memory stores that summarize past interactions \citep{zhong2024memorybank}, and token-level prompt compression \citep{jiang2023llmlingua}.
Their limitation is information loss: details that are not anticipated at summarization time, such as numbers, may disappear, and the summarizer typically does not see the question.
 
\paragraph{Retrieval and memory for assistants.}
Retrieval-augmented generation \citep{lewis2020rag} selects relevant text for each query.
Assistants apply it to dialogue memory by retrieving past events \citep{park2023generative} or managing memory tiers \citep{packer2023memgpt}.
Their limitation is retrieval failure when the request and the earlier statement share few words, which is typical for implicit preferences; the cost of reading or embedding the history is often left out.
 
\paragraph{Personalization and memory benchmarks.}
LaMP \citep{salemi2024lamp}, LoCoMo \citep{maharana2024locomo}, LongMemEval \citep{wu2025longmemeval}, PrefEval \citep{zhao2025prefeval} and PersonaMem \citep{jiang2025personamem,jiang2025personamemv2} test whether assistants use earlier interactions.
They mostly report accuracy of whole systems; they rarely isolate the effect of the context-construction step with references such as no context and the \textit{evidence only} oracle.
 
\paragraph{This work.}
We hold the judge, the question and the character budget fixed, vary only how the context is built, and add references that locate every result between ``no history'' and ``only the relevant history'', reporting what each context costs.
 
% Preamble: \usepackage{graphicx}
 
%\begin{figure*}[t]
%\centering
%\resizebox{0.97\textwidth}{!}{%
%\begin{tikzpicture}[
%  box/.style={draw, rounded corners=2pt, align=center, minimum height=9mm, font=\footnotesize, inner sep=3pt},
%  arr/.style={-{Stealth[length=2mm]}, thick},
%  lab/.style={font=\scriptsize\itshape}]
%\node[box] (hist) {Full history\\($\approx$157k chars)};
%\node[box, right=7mm of hist] (split) {Split into\\sentences};
%\node[box, right=7mm of split] (rank) {Embed and rank by\\similarity to the\\current request};
%\node[box, right=7mm of rank] (fill) {Fill budget $B$\\(1--5\% of history),\\restore order};
%\node[box, right=9mm of fill, minimum width=28mm] (reader) {Reader LLM\\+ current request\\$\rightarrow$ choose response};
%\node[box, below=7mm of rank] (sum) {LLM summary of the entire\\history ($\approx B$ chars)};
%\draw[arr] (hist) -- (split);
%\draw[arr] (split) -- (rank);
%\draw[arr] (rank) -- (fill);
%\draw[arr] (fill) -- (reader);
%\draw[arr] (hist.south) |- (sum.west);
%\draw[arr] (sum.east) -| (reader.south);
%\node[lab, above=1mm of rank] {(a) Relevance-filtered context (retrieval)};
%\node[lab, below=1mm of sum] {(b) Summary context};
%\end{tikzpicture}}
%\caption{Overview of the compared contexts. The history is turned into a compact context of $B$ characters by (a) sentence-level retrieval of the parts relevant to the current request or (b) an LLM summary of the whole history; the reader then chooses among candidate responses. Truncation (head or tail), the full history, the annotated\textit{evidence only}, and no context serve as comparisons.}
%\label{fig:pipeline}
%\end{figure*}
 
\section{Experimental Setup}
%\subsection{Setting}
 
\paragraph{Data.}
We use the text benchmark of PersonaMem-v2 \citep{jiang2025personamemv2} (32k-token histories; mean 156,594 characters in our sample).
An item has the user's current request, one correct personalized response, three distractors, and the annotated conversation snippet that reveals the relevant preference.
We use six preference types (stereotypical, anti-stereotypical, neutral, therapy background, health and medical conditions, ask-to-forget) and sample 30 items per type with a fixed seed and at most one item per persona, giving 180 independent four-way questions (chance 25\%).
We exclude the sensitive-information type because its evidence directly precedes the request.
The evidence snippet was located in the history for all 180 items.
 
\paragraph{Contexts.}
The \textit{full history} is the target, since a deployed assistant can always provide it.
\textit{Evidence only}, which gives the judge just the annotated snippet, is an oracle: it shows what perfect selection would achieve, but is unavailable in deployment.
\textit{No context} is the lower bound. \emph{Head} and \emph{tail slices} keep the first or last $B$ characters (or the first 25\% or 50\% of the history).
Our \textit{relevance-filtered context} splits every message into sentences (chunks of at most 250 characters), embeds them with \texttt{text-embedding-3-small} (\citet{openai2024embeddings}), ranks them by cosine similarity to the user's current request, and adds the top-ranked sentences until the budget $B$ is at least 90\% filled, restoring their original order; the budget thus acts as a stopping rule.
\emph{Summary} asks an LLM to summarize the \emph{entire} history to about $B$ characters under three headings (preferences, background, concerns); the summarizer does not see the question.
The main summarizer is \gpt{}; \haiku{} is an ablation.
Each context is built once per question and shown to both judges.
Budgets are $B=2{,}000$ and $8{,}000$ characters (about 1\% and 5\% of the history).
 
\paragraph{Judges and Evaluation.}
The judges are \gpt{} and \opus{}.
Given the context, the user's current request and the four candidate responses, the judge picks the response that best fits the user and answers with a single candidate, guessing when the context is inconclusive (Appendix~\ref{app:prompts}).
Unparseable answers are retried once and otherwise counted as incorrect (at most 1.1\% of answers in any condition).
We also report the \emph{gain recovered}, the share of the full history's improvement over no context that a given context scheme achieves (Table~\ref{tab:acc}).
Differences are tested with a paired bootstrap over questions \citep{efron1993bootstrap} (Appendix~\ref{app:full}).
 
\section{Results}
 
\begin{table*}[t]
\centering
\footnotesize
\setlength{\tabcolsep}{5pt}
\renewcommand{\arraystretch}{0.92}
\begin{tabular}{lrrlcccc}
\toprule
 & & & & \multicolumn{2}{c}{\gpt{}} & \multicolumn{2}{c}{\opus{}} \\
\cmidrule(lr){5-6}\cmidrule(lr){7-8}
Context & Chars & \% full & Built by & Acc. & Gain recovered & Acc. & Gain recovered \\
\midrule
No context & 30 & 0.0 & --- & 20.0 & 0.00 & 21.7 & 0.00 \\
Full history (target) & 156,594 & 100 & --- & 43.3 & 1.00 & 54.4 & 1.00 \\
Head slice, 8k & 7,473 & 4.8 & --- & 30.0 & 0.43 & 31.1 & 0.29 \\
Head slice, 50\% & 77,636 & 49.6 & --- & 36.1 & 0.69 & 48.3 & 0.81 \\
\midrule
Retrieval, 2k & 1,893 & 1.2 & embedding model & 37.2 & 0.74 & 46.7 & 0.76 \\
Retrieval, 8k & 7,331 & 4.7 & embedding model & 40.6 & 0.88 & \textbf{54.4} & \textbf{1.00} \\
Summary, 2k & 2,113 & 1.3 & LLM reading all & 36.7 & 0.72 & 41.7 & 0.61 \\
Summary, 8k & 8,649 & 5.5 & LLM reading all & \textbf{43.3} & \textbf{1.00} & 50.0 & 0.87 \\
\midrule
\textit{Evidence-only} (oracle) & 1,780 & 1.1 & annotation & 58.9 & 1.67 & 70.0 & 1.48 \\
\bottomrule
\end{tabular}
\caption{Accuracy (\%) and gain recovered ($n=180$; chance 25\%) with \gpt{} and \opus{} as judges ($n=180$; chance 25\%). Summaries are written by \gpt{}. Chars: mean characters given to the judge. Gain recovered: $(\mathrm{Acc}-\mathrm{Acc}_\text{none})/(\mathrm{Acc}_\text{full}-\mathrm{Acc}_\text{none})$. Bold: best non-oracle context. Details: Appendix~\ref{app:full}.}
\label{tab:acc}
\end{table*}
 
\begin{table*}[t]
\centering
\scriptsize
\setlength{\tabcolsep}{5pt}
\begin{tabular}{@{}lp{0.84\textwidth}@{}}
\toprule
\textbf{Request} & What's the most effective way to tackle early signs of aging like fine lines and uneven skin tone? \\\hline
\textbf{Correct response} & Since you prefer modern synthetic skincare over Ayurvedic remedies, a targeted routine with a gentle chemical exfoliant like glycolic acid, a peptide-infused moisturizer, and daily broad-spectrum SPF can be highly effective. \\
\midrule
\textbf{Retrieval, 8k} & [\ldots] USER: \textbf{I reached for the cool, luminous serum from my bathroom shelf\,---\,the one in the minimalist frosted bottle [\ldots] and smoothed it across my face.} [\ldots] USER: \textbf{It made me think about how, despite the pull of heritage remedies and nostalgia, I've always been drawn to the polished precision of modern formulations.} [\ldots] \\
\textbf{Summary, 8k} & \#\# Preferences [\ldots] The user is highly engaged with visual art, especially mixed media, layered compositions, tactile surfaces, found imagery, textiles, reclaimed materials, gold leaf, and the relationship between physical texture and memory [\ldots] \\
\bottomrule
\end{tabular}
\caption{Excerpts of the two compact contexts for one anti-stereotypical question. Bold: sentences that also appear in the annotated evidence. The retrieved sentences contain the user's preference for modern formulations over heritage remedies, whereas the summary, organized around the user's art interests, does not mention the serum or this preference. Both judges answered correctly with the retrieval context and incorrectly with the summary context (excerpts rebuilt with the same procedure).}
\label{tab:examples}
\end{table*}
 
\begin{figure*}[t]
\centering
\includegraphics[width=0.95\textwidth]{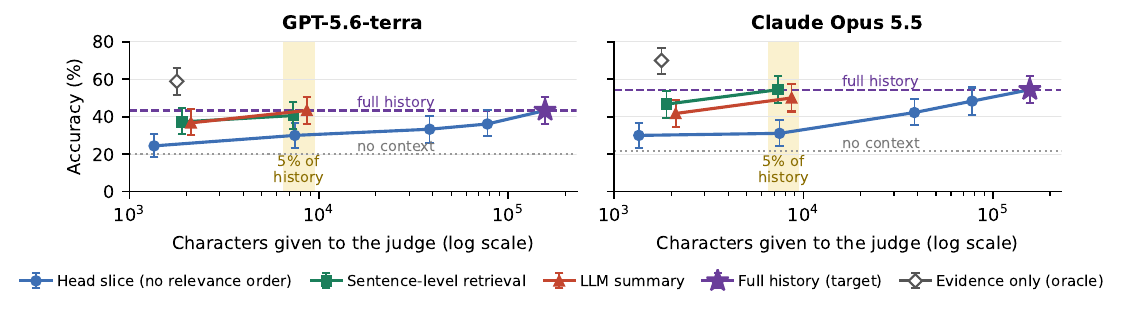}
\caption{Accuracy against the characters given to the judge. Retrieval and summaries with about 5\% of the history reach the full history (dashed), whereas head slices keep improving up to it. Error bars: 95\% bootstrap intervals.}
\label{fig:results}
\end{figure*}
 
\paragraph{Five percent of the history is on par with the full history.}
With about 5\% of the history (8k characters), the relevance-filtered context reaches 54.4\% with \opus{}, exactly the accuracy of the full history, and 40.6\% with \gpt{}, 2.8 points below it (Table~\ref{tab:acc}).
The LLM summary of the same length does the same: 43.3\% with \gpt{}, again exactly the full-history accuracy, and 50.0\% with \opus{}.
They recover 0.87--1.00 of the full-history gain, none differs significantly from it, and the judge reads 18--21 times fewer characters.
Even at 1\% of the history, retrieval recovers three quarters of the gain (0.74 and 0.76), about as much as the 40 times longer first half of the history (0.69 and 0.81).
 
\paragraph{Relevance ordering creates the stopping point.}
Without relevance ordering, more history keeps helping (Figure~\ref{fig:results}): head slices improve steadily up to the full history, and at 8k, where they contain the evidence for at most 6.7\% of the questions, they recover only 0.43 and 0.29 of the gain.
Once the history is ordered by relevance, the marginal gain from additional history vanishes after about 5\%: what matters is not how much the judge reads, but whether the relevant part comes first.

\paragraph{Retrieval or summary?}
Table~\ref{tab:examples} illustrates the difference on one question: retrieval keeps the user's own sentences verbatim, here the remark about preferring modern formulations, whereas the summary, written without seeing the request, condenses the history around the user's dominant topic, here art, and omits this detail. At equal budgets, the two compact contexts perform on par: their paired differences are small and not significant for both judges and budgets. They differ in what they require. A summary needs an LLM that reads the whole history, must be rewritten as the history grows, and depends on the summarizer: \gpt{} summaries outperformed \haiku{} summaries at 8k by $11.7$ and $10.0$ points, though the latter were also shorter. Retrieval needs only an embedding model, embeds each new message once, and uses the request to decide what to keep.

\paragraph{Consistency across judges and preference types.}
The judges order the contexts alike (Figure~\ref{fig:results}).
Across preference types (Appendix~\ref{app:types}), one pattern stands out: on ask-to-forget, where the user later asks the assistant to forget a preference, retrieval and summaries fall 16.7 and 33.3 points below the full history for both judges; a compact context keeps the preference but can drop the request to forget it.
On the other five types (150 questions), both compact contexts match or exceed the full history for both judges (0.0 to $+6.7$ points).
 
\paragraph{Headroom: the oracle beats the full history.}
The full history is not an upper bound: \textit{evidence only}, with 1.1\% of its text, is 15.6 points more accurate for both judges (58.9\% vs.\ 43.3\%; 70.0\% vs.\ 54.4\%), so irrelevant history actively hurts, consistent with position effects in long contexts \citep{liu2024lost}.
Retrieval leaves much of this headroom unused: by normalized string inclusion, at least one evidence sentence was retrieved for only 42\% of the questions at 2k and 61\% at 8k, and questions with retrieved evidence were answered 7 to 18 points more accurately in all four judge--budget settings.

\paragraph{Discussion: from \emph{how much} to \emph{which}.}
These results move the bottleneck from context length to context selection, with three directions.
First, a request rarely repeats the words of the preference it depends on (a morning routine may hinge on a remark about night shifts); having an LLM rewrite the request into the preferences it implies, or rerank a few dozen retrieved sentences, may target this gap at a small fraction of the cost of the full history.
Second, ask-to-forget shows that compact contexts treat the history as independent facts, although later statements can revoke earlier ones; memory should link such updates to what they modify.
Third, the two methods may complement each other: summaries did better when the correct response fits the user's typical profile (stereotypical preferences), and retrieval did better when the user departs from it (anti-stereotypical; Appendix~\ref{app:types}), so combining the two is worth examining.

\section{Conclusion}
A relevance-filtered context of 5\% of the conversation is on par with the full history for two judges at a twentieth of the input, while the right 1\% beats it by 15.6 points.
The open question for personalized memory is thus not how much history a model can read, but when to stop reading: an adaptive budget that halts where the marginal relevance of the next sentence drops, possibly learned from the judge's feedback, is a natural next step.
Memory benchmarks should likewise report equal-budget comparisons against no-context and oracle references.

\section*{Limitations}
\paragraph{Scope.} We evaluate one synthetic English benchmark (180 questions; six preference types of 30 questions each, so per-type results are descriptive), two judges, one embedding model and two budgets, with multiple-choice answers; results may differ for open-ended responses, real users or other models.
Although our earlier experiments also examined a Japanese setting (persona descriptions from Nemotron-Personas-Japan), they used a less controlled protocol and are preliminary (Appendix~\ref{app:earlier}), so whether the findings hold in other languages remains to be tested.
 
\paragraph{Statistical strength.} Each condition was run once, and intervals reflect only the sampling of questions; the full-history condition was run after the others.
No equivalence margin was fixed in advance, so ``is on par'' means that differences are small and not significant, not formal equivalence; comparisons are not corrected for multiple testing.
 
\paragraph{Comparability.} Retrieval sees the request whereas the summarizer does not, and \haiku{} summaries fell short of their budget.
 
\paragraph{References and diagnostics.} \textit{Evidence only} relies on the benchmark's annotation and is an empirical, not a guaranteed, upper bound; the retrieval diagnostic detects evidence by normalized string inclusion and is observational; cost is compared by characters read and models used, not measured prices or latency.

\section*{Ethical Considerations}
The benchmark is synthetic: personas, conversations and questions were generated by an LLM, and no real user data were collected or used.
Personalization from conversation history involves private information in real deployments; retrieving or summarizing a history also decides which personal details reach a model, and sensitive preferences (health, therapy, requests to forget information) need explicit consent and deletion mechanisms. Our ask-to-forget results show that a compact context can keep a preference the user asked to forget, so deletion requests must be applied to the stored history itself. We therefore ran no experiment on real users' data, and our results should not be read as an endorsement of storing or processing sensitive conversations.
Experiments used commercial APIs; we do not release personal information.
 
\bibliography{custom}
 
\appendix
 
\section{Prompts}
\label{app:prompts}
\paragraph{Judge.}
System prompt: ``You choose the response that best fits the user. If the context does not settle it, you must still choose the single most likely option. Reply with exactly one letter: a, b, c, or d.''
User prompt: the context (labelled as a conversation, possibly partial or a summary), the user's current request, the four candidate responses, and the instruction to reply with one letter.
On a parsing failure the prompt is repeated once with a reminder to reply with exactly one letter.
Output limits: 2,000 tokens for \opus{}; default for \gpt{}.
 
\paragraph{Summary.}
``Summarize the following conversation in about $B$ characters, organized under these headings: Preferences (food, travel, entertainment, lifestyle, etc.); Background (occupation, family, living situation, etc.); Concerns (health, worries, etc.). Keep concrete details (names, numbers, dates, habits).''
 
\section{All Conditions and Statistical Details}
\label{app:full}
Table~\ref{tab:allacc} lists all conditions.
Table~\ref{tab:ci} gives, for the main conditions, the paired difference from the full history with a 95\% percentile bootstrap interval over questions (4,000 resamples; \citealp{efron1993bootstrap}); a difference is called significant when its interval excludes zero.
No context, \textit{evidence only}, the 50\% head slice with \gpt{} and the 2k summary with \opus{} differ significantly from the full history; none of the 8k compact contexts does.
The paired differences between retrieval and \gpt{} summaries are not significant for either judge or budget (\opus{} at 8k: $+4.4$ [$-3.9$, $12.8$]).
Without any conversation, accuracy is not distinguishable from chance (two-sided binomial tests, $p=0.14$ and $0.34$).
In the retrieval diagnostic, the accuracy difference between questions with and without retrieved evidence is significant only for \gpt{} at 2k ($+17.6$ [$3.2$, $31.2$]).
 
\begin{table}[t]
\centering
\scriptsize
\setlength{\tabcolsep}{2pt}
\begin{tabular}{@{}lcc@{}}
\toprule
Context & \gpt{} & \opus{} \\
\midrule
No context & $-$23.3 [$-$30.0, $-$16.7] & $-$32.8 [$-$40.6, $-$25.0] \\
Evidence only & +15.6 [8.3, 23.3] & +15.6 [7.2, 23.9] \\
Head slice, 50\% & $-$7.2 [$-$13.3, $-$0.6] & $-$6.1 [$-$12.8, 0.0] \\
Retrieval, 2k & $-$6.1 [$-$13.3, 1.1] & $-$7.8 [$-$16.7, 1.1] \\
Retrieval, 8k & $-$2.8 [$-$10.0, 3.9] & 0.0 [$-$7.8, 7.8] \\
Summary, 2k & $-$6.7 [$-$14.4, 1.1] & $-$12.8 [$-$21.1, $-$4.4] \\
Summary, 8k & 0.0 [$-$7.8, 7.8] & $-$4.4 [$-$13.3, 3.9] \\
\bottomrule
\end{tabular}
\caption{Paired accuracy difference (points) from the full history, with 95\% bootstrap intervals over questions ($n=180$).}
\label{tab:ci}
\end{table}

\begin{table}[t]
\small
\centering
\footnotesize
\setlength{\tabcolsep}{2.5pt}
\begin{tabular}{@{}lrrcc@{}}
\toprule
Context & Chars & \% full & \shortstack{GPT-5.6-\\terra} & \shortstack{Claude\\Opus 5.5} \\
\midrule
No context & 30 & 0.0 & 20.0 & 21.7 \\
Evidence only (oracle) & 1,780 & 1.1 & 58.9 & 70.0 \\
Full history & 156,594 & 100 & 43.3 & 54.4 \\
\midrule
Head slice, 2k & 1,350 & 0.9 & 24.4 & 30.0 \\
Head slice, 8k & 7,473 & 4.8 & 30.0 & 31.1 \\
Head slice, 25\% & 38,490 & 24.6 & 33.3 & 42.2 \\
Head slice, 50\% & 77,636 & 49.6 & 36.1 & 48.3 \\
Tail slice, 2k & 1,716 & 1.1 & 25.6 & 30.6 \\
Tail slice, 8k & 6,923 & 4.4 & 24.4 & 33.9 \\
\midrule
Retrieval, 2k & 1,893 & 1.2 & 37.2 & 46.7 \\
Retrieval, 8k & 7,331 & 4.7 & 40.6 & \textbf{54.4} \\
Sum.(\gpt),2k & 2,113 & 1.3 & 36.7 & 41.7 \\
Sum.(\gpt),8k & 8,649 & 5.5 & \textbf{43.3} & 50.0 \\
Sum.(Haiku 4.5), 2k & 1,794 & 1.1 & 35.6 & 39.4 \\
Sum.(Haiku 4.5), 8k & 4,205 & 2.7 & 31.7 & 40.0 \\
\bottomrule
\end{tabular}
\caption{Accuracy (\%) of all conditions ($n=180$) with \gpt{} and \opus{} as judges. Bold: the highest value among non-oracle contexts.}
\label{tab:allacc}
\end{table}
 
\section{Consistency across Preference Types}
\label{app:types}
\begin{table*}[t]
\centering
\small
\setlength{\tabcolsep}{4pt}
\begin{tabular}{lrccccccccccc}
\toprule
 & & \multicolumn{5}{c}{\gpt{}} & & \multicolumn{5}{c}{\opus{}} \\
\cmidrule(lr){3-7}\cmidrule(lr){9-13}
Type & $n$ & None & Evid. & Full & Retr.\ 8k & Summ.\ 8k & & None & Evid. & Full & Retr.\ 8k & Summ.\ 8k \\
\midrule
Stereotypical & 30 & 16.7 & 63.3 & 46.7 & 53.3 & 66.7 & & 16.7 & 83.3 & 53.3 & 56.7 & 66.7 \\
Anti-stereotypical & 30 & 3.3 & 60.0 & 46.7 & 50.0 & 36.7 & & 3.3 & 70.0 & 56.7 & 50.0 & 33.3 \\
Neutral & 30 & 6.7 & 66.7 & 26.7 & 33.3 & 40.0 & & 10.0 & 80.0 & 40.0 & 46.7 & 60.0 \\
Therapy background & 30 & 33.3 & 46.7 & 46.7 & 30.0 & 56.7 & & 33.3 & 60.0 & 50.0 & 56.7 & 56.7 \\
Health and medical & 30 & 13.3 & 36.7 & 20.0 & 20.0 & 20.0 & & 10.0 & 50.0 & 36.7 & 43.3 & 26.7 \\
Ask-to-forget & 30 & 46.7 & 80.0 & 73.3 & 56.7 & 40.0 & & 56.7 & 76.7 & 90.0 & 73.3 & 56.7 \\
\midrule
All types & 180 & 20.0 & 58.9 & 43.3 & 40.6 & 43.3 & & 21.7 & 70.0 & 54.4 & 54.4 & 50.0 \\
Excl.\ ask-to-forget & 150 & 14.7 & 54.7 & 37.3 & 37.3 & 44.0 & & 14.7 & 68.7 & 47.3 & 50.7 & 48.7 \\
\bottomrule
\end{tabular}
\caption{Accuracy (\%) by preference type (30 questions each) for \gpt{} and \opus{}. None: no context; Evid.: \textit{evidence only}; Full: full history; Retr.\ 8k: retrieval with 8k characters; Summ.\ 8k: \gpt{} summary with 8k characters.}
\label{tab:bytype}
\end{table*}
 
Each type has 30 questions, so Table~\ref{tab:bytype} is descriptive.
The averages in Table~\ref{tab:acc} hide differences between types.
No-context accuracy ranges from 3.3\% (anti-stereotypical) to 56.7\% (ask-to-forget, \opus{}), below and above chance, so the near-chance average is a mixture.
\textit{Evidence only} is at least as accurate as the full history in all six types for \gpt{} (a tie for therapy background) and in five of six for \opus{}; the exception is ask-to-forget (76.7\% versus 90.0\%).
Retrieval with 8k characters is at or above the full history in four of six types for \gpt{} and four for \opus{}, and 16.7 points below it for ask-to-forget (both judges).
Summaries with 8k characters range from $-33.3$ to $+20.0$ points relative to the full history, and ask-to-forget is again the lowest ($-33.3$ for both judges).
Without ask-to-forget (150 questions), retrieval and summaries (8k) are between 0.0 and $+6.7$ points from the full history.
We looked at this breakdown after the main results, and each type has only 30 questions, so these observations are exploratory.
A possible reason for ask-to-forget is that retrieved sentences and summaries contain the preference the user later asked to forget, whereas the full history also contains the request itself; we did not test this.
 
\section{Preliminary Exploratory Experiments}
\label{app:earlier}
Before the main experiment we ran experiments on questions that an LLM generated from the same text: English conversations from PersonaMem-v2 (six domains; the questions were written in Japanese) and Japanese persona descriptions from Nemotron-Personas-Japan \citep{nvidia2025nemotronjp}.
We report only a small part of them, as preliminary results, for four reasons.
(1) No-context controls show that many questions could be answered without the history (chance 25\%): 50.5\% for \gpt{} on 200 questions about the English conversations, and 49.5\% (\gpt{}) and 51.5\% (\opus{}) on 200 Japanese questions. For Japanese, accuracy was 45.3\% and 48.0\% on the 179 numerical questions but 85.7\% and 81.0\% on the 21 non-numerical ones, which leave little room for a loss; the question files we could identify may not coincide exactly with the questions analysed below.
(2) The pipelines differed from the main experiment (answer format, output limits, answer parsing); in particular, \haiku{} answers were evaluated with a short output limit and a lenient parser, which can turn truncated or non-letter answers into errors.
(3) In the two result files with numerical and non-numerical questions that we re-checked, summaries and truncation were evaluated at different lengths, and the two methods were evaluated at the same length at one value only (100 characters for Japanese, 200 for English); in the file comparing domains they were evaluated at the same four lengths (200--2,000 characters). Averages over unmatched lengths mixed conditions, so we report only matched-length comparisons, with persona or question units as the unit of resampling; earlier significance tests counted repeated evaluations of the same question as independent and are not used.
(4) Other earlier findings (a sentence-level retrieval comparison in one domain, and judge dependence beyond what is shown below) have not been re-checked for these problems and are not reported.
 
The results we report are:
\begin{itemize}\setlength\itemsep{0pt}
\item \textbf{English conversations, numerical information.} At the only matched length (200 characters), summary-minus-truncation was $-11.4$ [$-24.4$, $+1.6$] points on 41 numerical questions and $-28.1$ [$-49.0$, $-6.3$] points on 16 non-numerical questions. The difference between the two went against a weakness specific to numbers and was not significant ($+16.7$ [$-8.7$, $+42.7$]; interaction $p=0.25$ with personas as clusters).
\item \textbf{English conversations, domains.} At the same lengths (200--2,000 characters), summaries were below truncation in the pooled data ($-15.1$ [$-23.4$, $-6.3$] points at 200 characters; $-32.7$ [$-39.1$, $-26.4$] at 2,000; 152 persona--domain units), and at 2,000 characters the interval excluded zero in five of six domains (Table~\ref{tab:domains}). The difference was large for \haiku{} ($-44.7$ [$-52.6$, $-37.5$] at 2,000) and small for \gpt{} ($-9.8$ [$-19.5$, $-1.2$]; 82 units); because the \haiku{} answers were evaluated with a short output limit and a lenient parser (item 2), this judge difference may partly reflect the evaluation.
\item \textbf{Japanese persona descriptions, numerical information.} Three experiments on numerical questions (80 questions each) gave summary-minus-truncation differences of $-3.7$ [$-8.7$, $+1.3$], $-35.0$ [$-46.3$, $-25.0$] and $-23.7$ [$-33.8$, $-15.0$] points (bootstrap over questions); %we did not re-check which lengths were compared in these files.
\item \textbf{Japanese persona descriptions, numerical versus non-numerical questions.} A small set of 20 numerical and 20 non-numerical questions cannot test whether a loss is specific to numbers: at the only matched length (100 characters) summaries were better for both question types (numerical $+20.0$ [$+5.0$, $+36.7$]; non-numerical $+58.3$ [$+38.3$, $+78.3$]), and the non-numerical questions could be answered without the persona description (item 1).
\end{itemize}
 
\begin{table}[t]
\centering
\small
\setlength{\tabcolsep}{4pt}
\begin{tabular}{lrc}
\toprule
Domain & Units & Summary $-$ truncation \\
\midrule
Entertainment & 17 & $-$29.4 [$-$44.1, $-$14.7] \\
Food & 17 & $-$35.3 [$-$50.0, $-$20.6] \\
Health & 70 & $-$38.1 [$-$49.5, $-$26.7] \\
Sports & 15 & $-$6.7 [$-$23.3, +10.0] \\
Technology & 16 & $-$31.2 [$-$46.9, $-$15.6] \\
Travel & 17 & $-$35.3 [$-$55.9, $-$14.7] \\
\midrule
All domains & 152 & $-$32.7 [$-$39.1, $-$26.4] \\
\bottomrule
\end{tabular}
\caption{Earlier experiments (English conversations, questions written in Japanese): accuracy difference (points) between summaries and truncation at the same length of 2,000 characters, with the judges \gpt{} and \haiku{} pooled, and 95\% bootstrap intervals over persona units.}
\label{tab:domains}
\end{table}
We therefore do not claim that summaries are specifically weak on numbers.
The main experiment controls the issues above (no-context and \textit{evidence only} references, forced choice with missing answers recorded, the same context shown to all judges, one question per persona); testing numerical details and domains with that protocol is future work.
 
\end{document}